\documentclass[10pt]{article} % For LaTeX2e

\usepackage[accepted]{rlj} % Should be uncommented for the camera-ready
\usepackage{amssymb}            % Defines common symbols like \mathbb R
\usepackage{mathtools}          % Extends amsmath, providing common math tools
\usepackage{mathrsfs}           % Enables \mathscr, which can work in cases that \mathcal does not
\usepackage{graphicx}           % For including images
\usepackage{subcaption}         % Allows for the use of subfigures and subcaptions
\usepackage[space]{grffile}     % For spaces in image names
\usepackage{url}                % For displaying URLs
\usepackage{lipsum}             % For placeholder text
\usepackage{booktabs}
\title{ATLAS: Adaptive Topological Learning with Abstract Successors for Continual Learning}

\setrunningtitle{ATLAS for Continual Learning}

\author{R. Blake Lawlor\textsuperscript{1}, Daniel S. Brown\textsuperscript{1}}

\emails{u1392237@utah.edu, daniel.s.brown@utah.edu}

\affiliations{
$^{1}$\textbf{Kahlert School of Computing, University of Utah}\\
}

\contribution{
    We introduce ATLAS, an architecture that achieves robust continual learning by structurally decoupling transition dynamics from the task's reward signal via a Grow When Required (GWR) topology and Successor Features.
    }
    {
    Prior transfer learning methods utilizing Successor Features typically require learning and storing new transformation matrices for every new task, whereas ATLAS utilizes a localized, task-agnostic topology.
    }

\contribution{
    We demonstrate that ATLAS achieves near-instantaneous adaptation and high sample efficiency in environments with dynamic goal relocations and shifting structural topologies.
    }
    {
    Standard on-policy and off-policy model-free baselines (PPO, SAC, DQN) fundamentally failed to bootstrap viable policies under the strict partial observability and sparse rewards of the tested environments.
    }

\contribution{
    We show that ATLAS exhibits robust topological retention in A-B-A task sequences. Where intermediate exploration reveals structural shortcuts, the agent successfully integrates these discoveries to formulate more highly optimized trajectories for previously mastered tasks.
    }
    {
    While the discrete topological mapping completely resists catastrophic forgetting, continuous-control low-level dynamics remain susceptible to localized forgetting, highlighting the exact boundary of our decoupled architecture.
    }

\contribution{
    We demonstrate that anchoring exact Successor Feature computation to a dynamically grown GWR topology enables highly efficient offline planning and rapid task adaptation.
    }
    {
   Standard continual learning methods that utilize Successor Features often rely on slow parametric updates. By directly inverting the transition matrix of our discrete topological graph, ATLAS drastically accelerates policy recovery without relying on vulnerable global dynamics models.
    }
\keywords{Continual Learning, Hierarchical Reinforcement Learning, Successor Features, Topological Memory, Catastrophic Forgetting, Model-Based Reinforcement Learning} % Your keywords

\summary{Contemporary model-free reinforcement learning algorithms can achieve very high performance, but have low sample efficiency and are not robust to changes in the environment. Model-based algorithms have much higher sample efficiency, but still fail when the environment shifts. This paper introduces Adaptive Topological Learning with Abstract Successors (ATLAS) to combat these challenges. ATLAS uses a Grow When Required network with Successor Features in order to achieve high sample efficiency while also robustly tackling catastrophic forgetting. We evaluate ATLAS in spatial navigation tasks, benchmarking its performance against common on-policy and off-policy algorithms. Our empirical results demonstrate that by structurally decoupling transition dynamics from the reward signal, ATLAS achieves near-instantaneous adaptation to new goals and can exhibit positive backward transfer, significantly outperforming baseline methods in non-stationary environments.
}

\begin{document}

% \makeCover  % Create the cover page
\maketitle  % Make the title section

\begin{abstract}
Contemporary model-free reinforcement learning algorithms can achieve very high performance, but have low sample efficiency and are not robust to changes in the environment. Model-based algorithms have much higher sample efficiency, but still fail when the environment shifts. This paper introduces Adaptive Topological Learning with Abstract Successors (ATLAS) to combat these challenges. ATLAS uses a Grow When Required network with Successor Features in order to achieve high sample efficiency while also robustly tackling catastrophic forgetting. We evaluate ATLAS in spatial navigation tasks, benchmarking its performance against common on-policy and off-policy algorithms. Our empirical results demonstrate that by structurally decoupling transition dynamics from the reward signal, ATLAS achieves near-instantaneous adaptation to new goals and can exhibit positive backward transfer, significantly outperforming baseline methods in non-stationary environments.
\end{abstract}

\section{Introduction}

A central challenge in continual reinforcement learning is balancing the stability-plasticity dilemma: enabling an agent to rapidly adapt to novel tasks (plasticity) while not overwriting previously acquired knowledge (stability). Current model-free approaches struggle with this, often suffering from catastrophic forgetting when goal locations or environmental dynamics shift. 

Model-based approaches improve sample efficiency by actively building and planning on environmental models, but standard methods still struggle with environmental shifts and catastrophic forgetting. To address this, we introduce a new model-based algorithm, Adaptive Topological Learning with Abstract Successors (ATLAS). ATLAS utilizes a Grow When Required (GWR) network \citep{marsland2002self} to learn the topology of the environment, and Successor Features (SF) \citep{barreto2017successor} to learn how to navigate the topology towards the goal. Our major contributions are as follows:
\begin{itemize}
    \item \textbf{Structural Decoupling for Continual Learning:} ATLAS naturally decouples the transition dynamics from the task's reward signal, making it robust to changes in the reward function.
    \item \textbf{Rapid Adaptation to Non-Stationarity:} In environments with dynamic goal relocation, ATLAS bypasses the need for global retraining, achieving fast recovery where standard model-free baselines completely fail.
    \item  \textbf{Topological Retention and Backward Transfer:}  We show that ATLAS robustly preserves spatial maps across sequential task shifts. Furthermore, when intermediate exploration reveals new structural shortcuts, the agent actively benefits by integrating these discoveries to formulate more highly optimized trajectories for previously mastered tasks.
    \item \textbf{Highly Efficient Offline Planning:} We utilize an offline planning mechanism ("dreaming") that directly computes global Successor Features across the discrete GWR graph. By bypassing slow parametric updates, this results in drastic improvements in sample efficiency over the baselines.
\end{itemize}

\section{Related Work}

\subsection{Continual Learning and Successor Features}
A primary challenge in lifelong reinforcement learning is catastrophic forgetting. Standard approaches mitigate this via weight regularization, such as Elastic Weight Consolidation \citep{kirkpatrick2017overcoming}, or dynamic architectures like Progressive Neural Networks \citep{rusu2016progressive}. However, these methods often require explicit task-boundary signals and scale poorly across lifelong horizons due to continuous architectural bloat. Transfer learning approaches leveraging Successor Features (SF) \citep{barreto2017successor, zhang2017deep} offer an alternative by decoupling environment dynamics from rewards, allowing for rapid task adaptation. Yet, they still require learning and storing additional transformation matrices for every new task encountered. ATLAS resolves these scaling limitations by mapping the state-space into a discrete, self-organizing Grow When Required (GWR) topology. This inherently localizes feature representations, enabling task-agnostic continual learning without explicit task-switching signals for memory consolidation or task-specific feature transformations.

\subsection{Topological Memory and Offline Planning}
To bridge planning and reinforcement learning in long-horizon tasks, methods such as Search on the Replay Buffer \citep{eysenbach2019search} and Semi-Parametric Topological Memory \citep{savinov2018semi} construct navigation graphs from past experiences. Because these memory structures evaluate global pairwise distances or grow linearly with exploration time, they are computationally prohibitive for large, lifelong environments. Concurrently, model-based planners like World Models \citep{ha2018world} and Dreamer \citep{hafner2019dream} achieve high sample efficiency via latent offline planning but remain highly susceptible to catastrophic forgetting due to their reliance on global recurrent dynamics models. ATLAS retains the sample efficiency of latent state compression but replaces global RNNs and exhaustive replay buffers with its dynamically compressed GWR topology. By anchoring offline value propagation directly to this sparse, localized graph structure, ATLAS structurally insulates prior knowledge from catastrophic interference while maintaining a bounded memory footprint.

\subsection{Biological Inspiration: Hippocampal Place Cells}

ATLAS was inspired by hippocampal place cells in mammals. At a high level, these are cells located in CA1 that activate only when the animal is in a certain location of the environment \citep{o1978hippocampus,o1971hippocampus}. Our GWR structure is designed to functionally mimic these place cells to anchor topological memory, while Successor Representations are postulated to be the exact biological mechanism for predictive navigation in mammals \citep{stachenfeld2017hippocampus}. While we use Successor Features in ATLAS, it is a computationally useful abstraction of Successor Representations.

\section{The ATLAS Architecture}

\subsection{Topological Memory via Grow When Required}

Grow When Required (GWR), originally introduced by \citet{marsland2002self}, is a self-organizing topological network designed for continual, online learning. ATLAS implements a modified variant of this architecture. Given a new input state, the network first determines the best-matching unit (BMU), defined as the existing topological node with the minimum Euclidean distance $\rho$ to the input. The activation $a_s$ of this BMU is modeled as an exponential decay of this distance: $a_s = e^{-\rho}$.

If the activation falls below a predefined insertion threshold ($a_s < a_t$), the input is classified as sufficiently novel, prompting the initialization of a new node that is immediately linked to the preceding node. Conversely, if the input is not sufficiently novel, no new node is created. Instead, the agent simply encodes the temporal transition by generating an edge from the previous best-matching unit (BMU) to the current BMU, leaving the existing weights entirely unmodified. 

Furthermore, whereas traditional GWR formulations dynamically adjust node weights, employ pruning mechanisms, and utilize undirected graphs, ATLAS strictly enforces static weights, a monotonically growing topology, and directed edges. By permanently freezing node weights upon initialization and preserving the asymmetric reachability of the environment, ATLAS maintains the strict representational stability required for continual learning. 

\subsection{Global Routing with Successor Features}
To navigate the learned topology, we utilize Successor Features (SF), as introduced by \citet{barreto2017successor}. Beyond simply storing topological state embeddings, each node in the GWR network also maintains a continuous successor feature vector, $\psi^\pi(s, a)$. This vector is defined as the expected discounted sum of future state features $\phi$ encountered when following a policy $\pi$ from an initial state-action pair $(s, a)$:

$$\psi^\pi(s, a) = \mathbb{E}_\pi \left[ \sum_{t=0}^{\infty} \gamma^t \phi(s_t, a_t, s_{t+1}) \mid s_0=s, a_0=a \right]$$

The core assumption of this framework is that the immediate reward for any given task can be approximated by a linear combination of these features and a task-specific weight vector \textbf{w}, such that:

$$r(s, a, s') \approx \phi(s, a, s')^\top \mathbf{w}$$

We can then derive the Q-value function as:
$$Q^\pi(s, a) \approx \psi^\pi(s, a)^\top \mathbf{w}$$

By using Successor Features, the dynamics of the environment, $\psi^\pi(s, a)$, are split from the reward signal, \textbf{w}. If the task changes, the agent no longer suffers from catastrophic forgetting because it does not need to relearn the environment dynamics, it simply updates the task-specific weight vector.

To ensure that the agent can respond quickly to reward and environment changes, we introduce an offline "dreaming" phase to the SF learning. Every $n$ episodes, we solve for the global SF matrix directly to bypass the slow convergence of incremental Temporal Difference (TD) updates:

$$\mathbf{\Psi} = (\mathbf{I} - \gamma \mathbf{T})^{-1}$$

where $\Psi$ is the successor matrix, $\gamma$ is the discount factor, and $\mathbf{T}$ is the transition matrix derived from the directed edges in the GWR.

\subsection{The High-Level Controller}
The High-Level Controller (HLC) outputs subgoals for the Low-Level Controller (LLC) to route to. For each node in the topology, the HLC assigns it a score that balances an intrinsic curiosity with an extrinsic reward signal. Each node starts with a base curiosity formulated as: $C(v_n)=\frac{1}{\sqrt{v_n + 1}}$, where $v_n$ is the number of times node $n$ has been visited. To encourage the agent to explore unmapped areas, each node receives a frontier bonus inversely proportional to its degree:

$$\mathcal{F}(d_n) = \begin{cases} 5.0, & \text{if } d_n \le 1 \\ 2.0, & \text{if } d_n = 2 \\ 1.0, & \text{if } d_n > 2 \end{cases}$$

While the count bonus ensures general state visitation, the frontier bonus helps encourage boundary expansion. Without the frontier bonus, an agent could become trapped traversing a uniformly explored interior rather than pushing into new topological space. The final value is calculated as:

$$S(n) = \boldsymbol{\psi}_n^\top \mathbf{w} + \beta \Big( \mathcal{C}(v_n) \cdot \mathcal{F}(d_n) \Big)$$

with $\beta$:

$$\beta = \begin{cases} 0, & \text{if } \max_{i} (\boldsymbol{\psi}_i^\top \mathbf{w}) \ge \tau \\ \beta_0, & \text{otherwise} \end{cases}$$

where $\tau$ is a hyperparameter that defines when to trust the graph and end exploration, and $\beta_0$ is a hyperparameter that controls the magnitude of the intrinsic curiosity drive.

Once a target node is selected by maximizing $S$, the HLC must route the agent through the topology. While it is theoretically possible to navigate toward extrinsic goals by greedily following the Successor Feature value gradient, this approach is incompatible with intrinsic exploration; because SF values are policy-conditioned, they do not natively encode gradients towards purely curiosity-driven targets. Consequently, to maintain algorithmic uniformity and simplicity, ATLAS universally employs a Breadth-First Search (BFS) over the directed GWR graph. This extracts the shortest topological path to any selected subgoal, regardless of whether the node was prioritized for its reward or its novelty. The HLC returns the first node on this path to the LLC.

\subsection{The Low-Level Controller}
The Low-Level Controller (LLC) operates as a goal-conditioned inverse dynamics model, parameterized by a small, fully connected neural network. Its sole responsibility is local navigation. Let $\mathcal{S}$ be the environment state space and $\mathcal{A}$ the action space. At timestep $t$, the LLC takes as input the current state $s_t \in \mathcal{S}$ and a local target state $g \in \mathcal{S}$ (the coordinates of the subgoal provided by the HLC), and outputs the action $a_t \in \mathcal{A}$ required to traverse the gap.

Rather than relying on standard temporal difference learning, the LLC is trained strictly via supervised learning as a goal-conditioned policy trained via inverse dynamics. We sample trajectory windows of maximum size $K$ from an episodic working memory buffer. By training the network to minimize the error between its predicted action and the actual executed action given a state $s_t$ and an achieved future state $s_{t+k}$ (where $k \in [1, K]$), the LLC learns a robust, generalized inverse model.

Crucially, this multi-horizon training formulation ensures the LLC can smoothly navigate varying topological distances. Because the network learns vector and trajectory scaling natively across diverse time horizons, it reliably drives the agent toward GWR subgoals even when traversing the intermediate space requires complex, multi-step action sequences.

\subsection{State Representation Decoupling}
\begin{figure}
    \centering
    \includegraphics[width=0.8\linewidth]{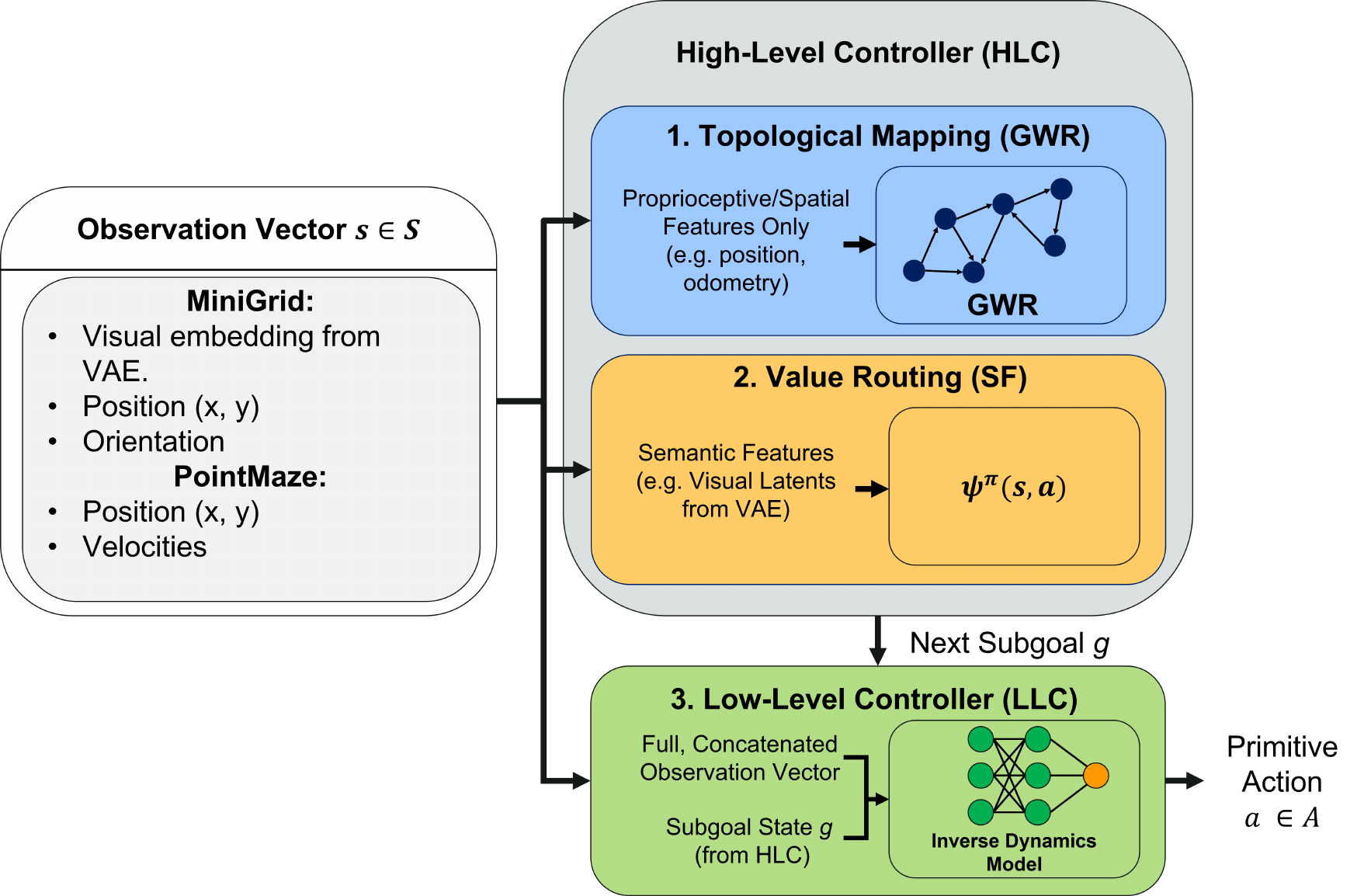}
    \caption{The ATLAS architecture. The observation gets split such that the proprioceptive and spatial features get sent into the GWR, the semantic features go to the SF, and the LLC gets the entire observation. The GWR and SF (HLC) compute the node with the highest score and send it to the LLC. The LLC takes the observation and the next subgoal and outputs the first action needed to reach the subgoal.}
    \label{fig:atlas}
\end{figure}

Unlike standard monolithic reinforcement learning models that entangle spatial mapping with value estimation, the core innovation of ATLAS lies in explicitly decoupling these components. By utilizing the GWR network to maintain a stable, task-agnostic spatial topology and Successor Features to route semantic value, the architecture ensures that the agent only needs to update its reward vector when objectives shift, entirely bypassing catastrophic forgetting. The complete ATLAS structure is illustrated in Figure \ref{fig:atlas}.

To ensure topological stability across diverse domains, ATLAS enforces a strict informational bottleneck. The GWR receives only low-dimensional spatial features to prevent visual over-segmentation, the SF processes task-relevant semantic embeddings to evaluate reward, and the LLC receives the full concatenated state to execute precise local kinematics.

\section{Results}

We evaluate ATLAS in two distinct domains: PointMaze Medium \citep{gymnasium_robotics2023} and MiniGrid Four Rooms \citep{chevalierboisvert2023minigrid}. By validating the architecture across both continuous and discrete control tasks, we demonstrate its capacity to generalize across fundamentally different environmental mechanics.

To strictly isolate our analysis of adaptation and positive backward transfer during A-B-A task transitions, explicit exploration resets ($\epsilon$-decay and entropy schedules) were applied at task boundaries for both ATLAS and the discrete baseline algorithms in the MiniGrid evaluations. Conversely, the continuous control baselines in PointMaze relied exclusively on native mechanics (e.g., SAC's automatic entropy tuning) to prevent artificial disruption to their action distributions (see Appendix \ref{app:exploration} for details and boundary-agnostic theoretical extensions).

\subsection{Experimental Setup}
Both environments utilize an A-B-A task sequence to evaluate resistance to catastrophic forgetting and capacity for positive backward transfer. Across all plotted evaluations, shaded regions denote $\pm 1$ standard deviation over 5 random seeds. Visuals for both environments are found in Appendix \ref{app:configs}.

\textbf{PointMaze (Continuous):} A $1 \times 10^6$ step sequence where a sparse-reward goal relocates from the bottom-right (Phase A), to the top-right at step 333,333 (Phase B), and back to the bottom-right at step 666,666 (Phase A2).

\textbf{MiniGrid (Discrete):} A 300,000 step sequence evaluating simultaneous shifts in rewards and transition dynamics. In Phase A, the top-right room is locked. In Phase B (step 100,000), the room unlocks and the goal moves inside. In Phase A2 (step 200,000), the goal returns to its original position, but the room remains unlocked, offering a significantly shorter topological route. 

% ---------------------------------------------------------
% THE COMBINED FIGURE BLOCK GOES HERE
% ---------------------------------------------------------
\begin{figure*}[t] 
    \centering
    \begin{subfigure}[b]{0.75\textwidth}
        \centering
        \includegraphics[width=\textwidth]{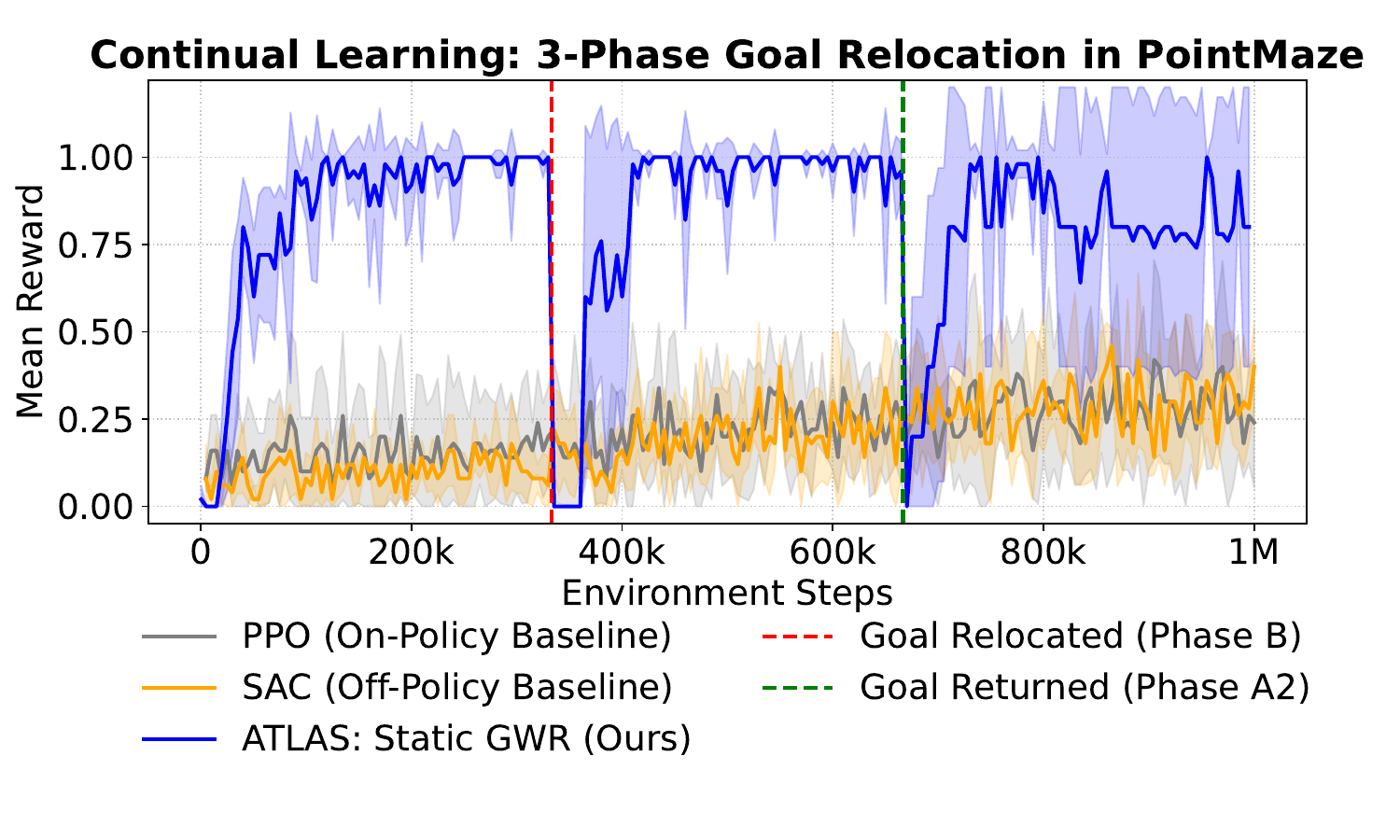}
        \vspace{-10mm}
        \caption{PointMaze Goal Relocation}
        \label{fig:pointmaze}
    \end{subfigure}
    \begin{subfigure}[b]{0.75\textwidth}
        \centering
        \includegraphics[width=\textwidth]{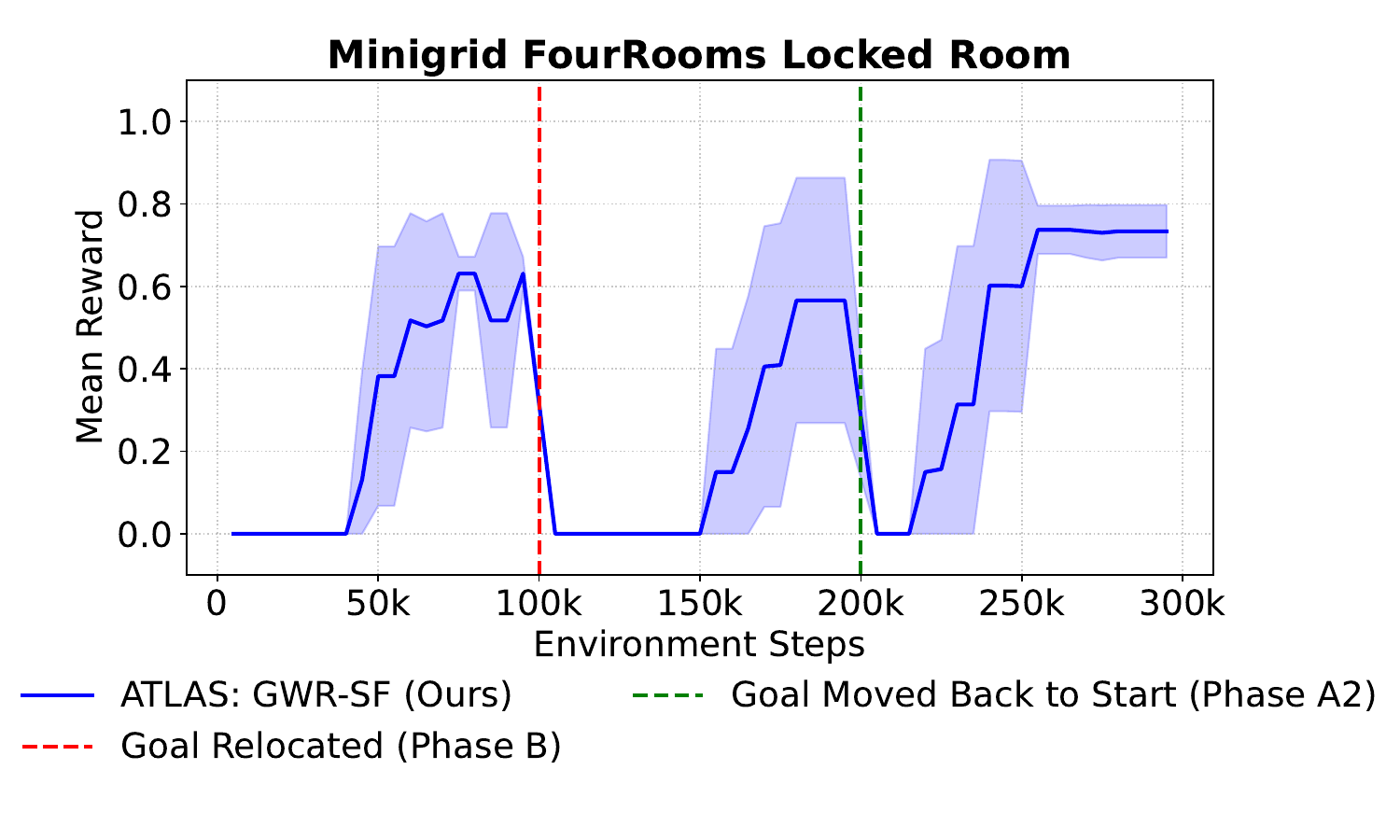}
        \caption{MiniGrid Dynamic Topology}
        \label{fig:locked}
    \end{subfigure}
    \vspace{-2mm}
    \caption{Continual learning performance across discrete and continuous environments. (a) ATLAS rapidly adapts to a non-stationary goal in continuous space, whereas baselines fail to bootstrap, reamining effectively random. (b) When a previously locked shortcut is opened, ATLAS exhibits positive backward transfer, actively optimizing its original trajectory.}
    \label{fig:all_results}
    \vspace{-3mm}
\end{figure*}
% ---------------------------------------------------------

\subsection{Rapid Adaptation to Non-Stationarity}
Across both continuous and discrete domains, standard model-free baselines (SAC, PPO, and DQN) fundamentally failed to bootstrap viable policies for the initial tasks in the relatively limited allocated timesteps. Given more time, these algorithms would likely achieve rewarding policies, but they are far too sample-inefficient to do so in these short experiments (Figure \ref{fig:all_results}a, Appendix \ref{app:experiments}). Consequently, evaluating their capacity for catastrophic forgetting during task shifts is moot; their behavior remains effectively random.

Conversely, ATLAS demonstrates high sample efficiency, rapidly converging on stable policies in Phase A. Following the Phase B goal shifts, ATLAS leverages its decoupled transition dynamics, requiring merely a few thousand timesteps of localized exploration to identify the new targets and restabilize. It is worth noting that, when the task shifts, ATLAS receives no signal about where the goal relocated to, rather it must re-explore the environment. By bypassing the need for global parametric retraining, ATLAS achieves near-instantaneous recovery.

\subsection{Topological Recovery and Backward Transfer}
When the objectives revert to their original locations (Phase A2), ATLAS leverages its preserved topological map to recover rapidly, though the nature of this recovery differs by domain.

\begin{figure}
    \centering
    \includegraphics[width=0.5\linewidth]{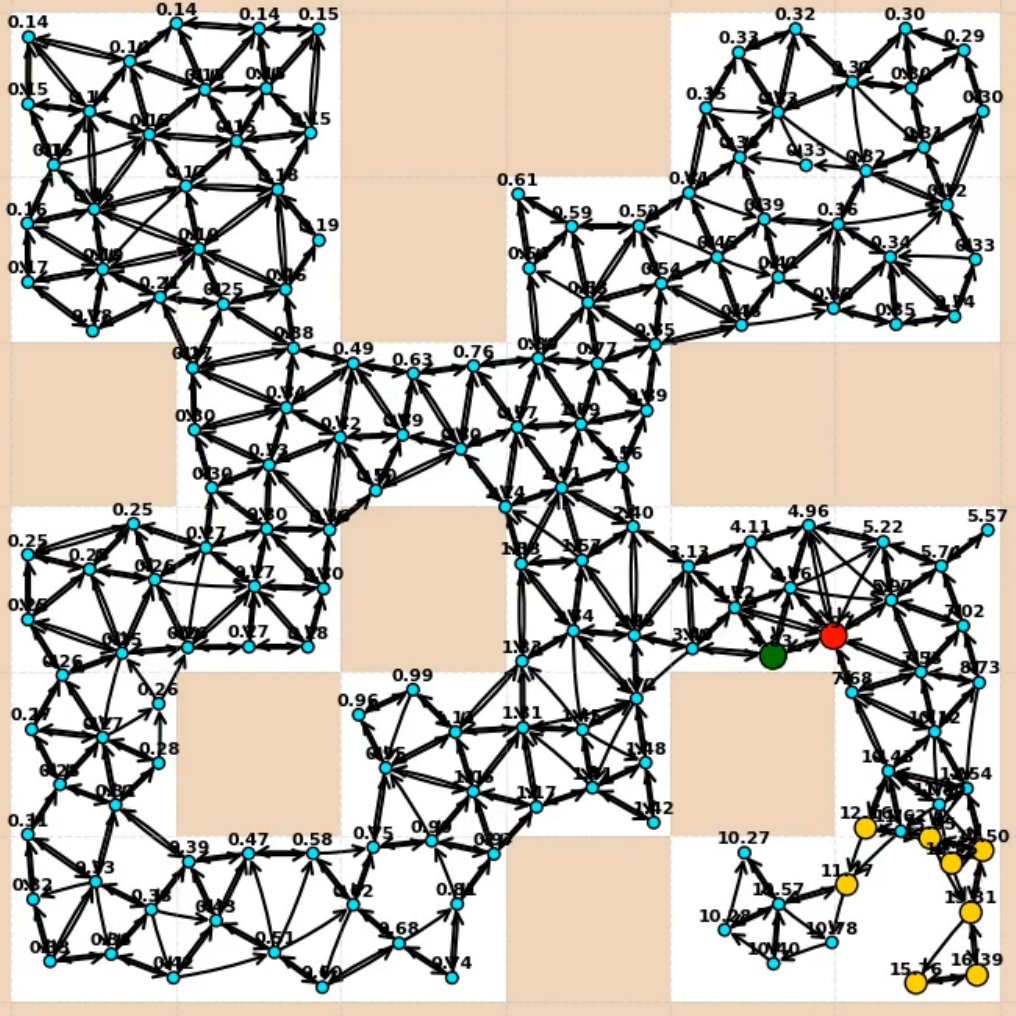}
    \caption{Example of built topology for the PointMaze Medium environment. Each node has their SF value, which creates a gradient towards the goal nodes (denoted in yellow). The green node is where the agent currently is, and the red node is the subgoal. There are multiple goal nodes as PointMaze counts any position within a threshold of the actual goal. We initially experimented with a lower $a_t$, but found it would create edges through walls. Investigating dynamic edge deletion remains an avenue for future work.}
    \label{fig:example}
\end{figure}

In PointMaze, the High-Level Controller robustly retains the topological path to the original goal, routing the agent to the target slightly faster than in Phase A. However, the recovered policy exhibits significant episodic variance and a lower asymptotic mean. This instability explicitly highlights the decoupled nature of the architecture: while the HLC remembers the route, the Low-Level Controller suffers from partial catastrophic forgetting of the specific continuous-control dynamics required to navigate it smoothly. An image of ATLAS's topology in PointMaze can be seen in Figure ~\ref{fig:example}.

In MiniGrid, ATLAS demonstrates true positive backward transfer. Rather than rigidly adhering to the safe, known route established in Phase A, the agent successfully integrates the newly discovered topological shortcut from Phase B into its global map. By utilizing this intermediate knowledge, ATLAS yields a Phase A2 policy with a substantially higher asymptotic mean and lower variance than the original training phase, actively improving its execution of the original task.

\subsection{Ablation Study: Topological Stability}

We empirically validate the necessity of ATLAS's structural constraints through an ablation study in the continuous PointMaze Medium environment. To isolate the impact of our modifications, we evaluated a baseline agent reverted to standard GWR mechanics—specifically, utilizing dynamic node weight adjustments, age-based edge pruning, and undirected graphs. 

\begin{figure}[ht]
    \centering
    \includegraphics[width=0.8\linewidth]{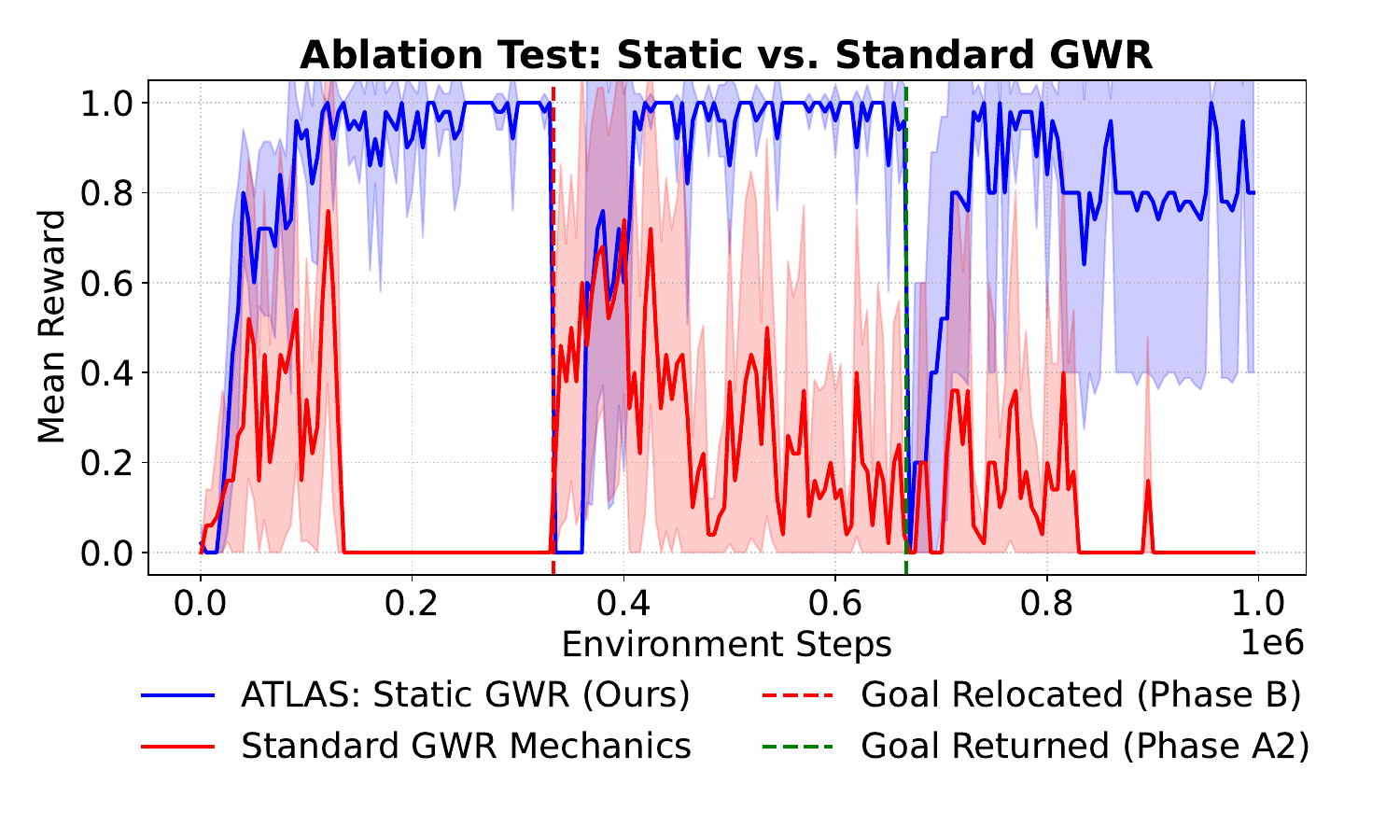}
    \caption{Ablation study in the continuous PointMaze Medium environment evaluating the necessity of ATLAS's structural constraints. Standard GWR with shifting weights and undirected edges fails to consistently reach the goal before it is even relocated, as dynamic weights pull critical nodes into impassible walls. ATLAS maintains its topology across all phases, demonstrating the strict necessity of static weights and directed edges for successful navigation.}
    \label{fig:ablation}
\end{figure}

As demonstrated in Figure \ref{fig:ablation}, the standard GWR architecture experiences catastrophic topological collapse. Because standard continuous interpolation dynamically adjusts node coordinates during traversal, the agent's spatial map shatters as critical nodes are incrementally shifted into impassable walls or pulled out of viable navigational corridors. Consequently, the baseline agent fails to consistently reach the goal even during the initial mapping phase. 

In contrast, by permanently freezing spatial coordinates upon initialization and enforcing directed edges, ATLAS successfully maintains its topological integrity, demonstrating the strict necessity of these mechanisms for successful navigation in non-stationary environments.

\section{Discussion}
These results demonstrate that ATLAS is a robust framework for continual learning in both discrete and continuous spatial navigation. Building on this foundation, several exciting avenues for future research emerge: (1) further fortifying the Low-Level Controller against localized forgetting by integrating advanced regularization techniques such as Elastic Weight Consolidation \citep{kirkpatrick2017overcoming} or Mixture of Experts architectures \citep{jacobs1991adaptive, shazeer2017outrageously}; (2) extending the topological mapping to highly stochastic environments with continuously moving obstacles, where dynamic edge deletion mechanisms could enable deployment in complex, real-world robotics applications; (3) abstracting the core architecture into a generalized semantic engine, where nodes encode conceptual representations rather than physical coordinates, potentially unlocking broad generalization across diverse, non-spatial tasks; and (4) benchmarking ATLAS against broader model-based baselines to further highlight its comparative sample efficiency and structural advantages. Ultimately, by isolating the "where" (topology) from the "why" (reward), ATLAS provides a highly sample-efficient, continual learning foundation that avoids the pitfalls of black-box model-free approaches.

\bibliography{main}
\bibliographystyle{rlj}

%%%%%%%%%%%%%%%%%%%%%%%%%%%%%%%%%%%%%%%%%%%%%%%%%%%%%%%%%%%%%%%%
% AUTHOR: If your paper has no supplementary materials, you may 
%         comment out the line below, which creates the title for
%         the supplementary materials.
%%%%%%%%%%%%%%%%%%%%%%%%%%%%%%%%%%%%%%%%%%%%%%%%%%%%%%%%%%%%%%%%
\beginSupplementaryMaterials

\section{Implementation Details and Hyperparameters}
\label{app:implementation}

\subsection{Network Architectures}
\label{app:networks}

% Describe the sizes of your networks here so people can reproduce your work.
The Low-Level Controller (LLC) was parameterized as a fully connected Multi-Layer Perceptron (MLP) consisting of 2 hidden layers with 256 units each, utilizing ReLU activations. To accommodate the differing action spaces of the test environments, the output layer was modified accordingly. For the discrete Minigrid tasks, the final layer outputs the raw logits which we take the argmax of. For the continuous PointMaze tasks, the final layer unitizes a tanh function to strictly bound the continuous directional velocity outputs to the environments $[-1, 1]$ action limits.

For the VAE, we optimized the network using Binary Cross-Entropy (BCE) loss, which empirically provided the highest reconstruction fidelity and latent space stability compared to MSE and L1.

The encoder features three convolutional layers (4x4 kernel, stride 2, padding 1) with SiLU activations \citep{elfwing2018sigmoid}. The resulting feature maps are flattened, layer-normalized, and processed by two parallel linear layers to output the latent mean and log-variance.

The symmetric decoder unflattens the latent vector via a linear layer, followed by three transposed convolutional layers with matching hyperparameters. To ensure numerical stability, the decoder omits a final activation function, computing the BCE loss directly from the output logits.

\subsection{Hyperparameter Specifications}
\label{app:hyperparameters}
Table \ref{tab:hyperparameters} details the hyperparameters utilized across both the PointMaze and MiniGrid environments.

\begin{table}[htbp]
\centering
\caption{Hyperparameters for ATLAS across all evaluated environments.}
\label{tab:hyperparameters}
\begin{tabular}{@{}llcc@{}} % The @{} removes the invisible vertical padding at the edges
\toprule
\textbf{Component} & \textbf{Hyperparameter} & \textbf{PointMaze} & \textbf{MiniGrid} \\ 
\midrule

\textbf{Variational Autoencoder} & Latent Dimension & - & 32
\\
                                & KL Penalty ($\beta$) & - & 0.01
\\
                                & Learning Rate & - & $5 \times10^{-4}$
\\
                                & Batch Size & - & 64
\\
\midrule
\textbf{Topological Mapping (GWR)} & Insertion Threshold ($a_t$) & 0.7 & 0.8 \\

& BMU Learning Rate  & 0.0 & 0.05* \\

\midrule

\textbf{High-Level Controller (SF)} & SF Dream Frequency ($n$ episodes) & 30 & 30 \\
                                   & Trust Threshold ($\tau$) & 0.5 & 0.5 \\
                                   & Discount factor ($\gamma$) & 0.99 & 0.99 \\
                                   & Intrinsic Curiosity Drive ($\beta_0$) & 0.5 & 0.5
                                   \\

\midrule

\textbf{Low-Level Controller (LLC)} & Learning Rate & $3 \times 10^{-4}$ & $1 \times 10^{-3}$ \\
                                   & Batch Size & 256 & 128 \\
                                   & Context Window ($K$) & 5 & 5 \\

\bottomrule
\end{tabular}
\vspace{1ex} % Adds a tiny bit of vertical space below the table
\raggedright \footnotesize 
\textit{Note: Only the visuals receive this update, the orientation and odometry weights have a learning rate of 0.0.}

\end{table}

For the continuous PointMaze environment, we evaluated both Proximal Policy Optimization (PPO) and Soft Actor-Critic (SAC) (Tables 2 and 3). For the discrete MiniGrid tasks detailed in Appendix \ref{app:experiments}, we evaluated RecurrentPPO to provide the agent with necessary LSTM memory (Table 2). Additionally, we evaluated a Deep Q-Network (DQN). To handle MiniGrid's partial observability without recurrence, the DQN baseline utilized Frame Stacking paired with a standard MLP policy (Table 4).

\begin{table}[htbp]
\centering
\caption{Hyperparameters for baseline algorithms. Note that due to the partial observability of the MiniGrid environment, the baseline was upgraded to RecurrentPPO utilizing an LSTM memory architecture to ensure a fair comparison.}
\label{tab:baseline_hyperparams}
\begin{tabular}{@{}llcc@{}} 
\toprule
\textbf{Algorithm} & \textbf{Hyperparameter} & \textbf{PointMaze} & \textbf{MiniGrid} \\ 
\midrule

\textbf{Architecture} & Algorithm Variant & Standard PPO & RecurrentPPO \\
                      & Policy Type & \texttt{MlpPolicy} & \texttt{MlpLstmPolicy} \\

\midrule

\textbf{Optimization} & Learning Rate & $3 \times 10^{-4}$ & Linear Decay ($3 \times 10^{-4} \to 0$) \\
                      & Rollout Steps (\texttt{n\_steps}) & 2048 & 4096 \\
                      & Batch Size & 64 & 128* \\
                      & Number of Epochs & 10* & 10* \\

\midrule

\textbf{Loss Function} & Entropy Coef. (\texttt{ent\_coef}) & 0.01 & Linear Decay ($0.1 \to 0.005$) \\
                       & Value Function Coef. & 0.5* & 0.5* \\
                       & Clip Range & 0.2* & 0.2* \\
                       & GAE Lambda ($\lambda$) & 0.95* & 0.95* \\
                       & Discount Factor ($\gamma$) & 0.99* & 0.99* \\
                       & Max Gradient Norm & 0.5* & 0.5* \\

\bottomrule
\end{tabular}

\vspace{1ex}
\raggedright \footnotesize 
\textit{* Note: Values marked with an asterisk indicate the default Stable Baselines3 configurations, as they were not explicitly overridden in the experimental setup.}

\end{table}

\begin{table}[htbp]
\centering
\caption{Hyperparameters for the Soft Actor-Critic (SAC) baseline evaluated in the continuous PointMaze environment.}
\label{tab:sac_hyperparams}
\begin{tabular}{@{}llc@{}} 
\toprule
\textbf{Category} & \textbf{Hyperparameter} & \textbf{Value} \\ 
\midrule

\textbf{Architecture} & Policy Type & \texttt{MlpPolicy} \\

\midrule

\textbf{Optimization} & Learning Rate & $3 \times 10^{-4}$ \\
                      & Batch Size & 256 \\
                      & Replay Buffer Capacity & 500,000 \\
                      & Learning Starts (steps) & 10,000 \\

\midrule

\textbf{Q-Learning \& Entropy} & Discount Factor ($\gamma$) & 0.99 \\
                               & Target Smoothing ($\tau$) & 0.005 \\
                               & Target Update Interval & 1 \\
                               & Entropic Temperature ($\alpha$) & Auto (Learnable) \\

\bottomrule
\end{tabular}
\end{table}

\begin{table}[htbp]
\centering
\caption{Hyperparameters for the Deep Q-Network (DQN) baseline evaluated in the discrete MiniGrid environment. To overcome the partial observability of the environment without relying on recurrent memory, the baseline utilized Frame Stacking across recent observations.}
\label{tab:dqn_hyperparams}
\begin{tabular}{@{}llc@{}} 
\toprule
\textbf{Category} & \textbf{Hyperparameter} & \textbf{Value} \\ 
\midrule

\textbf{Architecture} & Policy Type & \texttt{MlpPolicy} (with Frame Stacking) \\

\midrule

\textbf{Optimization} & Learning Rate & $1 \times 10^{-4}$ \\
                      & Batch Size & 128 \\
                      & Replay Buffer Capacity & 100,000 \\
                      & Learning Starts (steps) & 10,000 \\
                      & Train Frequency & 4 \\

\midrule

\textbf{Q-Learning} & Discount Factor ($\gamma$) & 0.99 \\
                    & Target Update Interval & 1,000 \\
                    & Target Smoothing ($\tau$) & 1.0 (Hard Update) \\

\midrule

\textbf{Exploration} & Initial $\epsilon$ & 0.5 \\
                     & Final $\epsilon$ & 0.01 \\
                     & Decay Fraction & 0.2 (First 20\% of training) \\

\bottomrule
\end{tabular}
\end{table}

\clearpage

\subsection{Exploration Strategy}
\label{app:exploration}

To ensure sufficient interaction with the environment during the initial topological mapping phase, ATLAS utilized an $\epsilon$-greedy exploration strategy in conjunction with the intrinsic curiosity bonus calculated by the High-Level Controller. 

During the exploration phase of each task, the agent selected a random action from the action space $\mathcal{A}$ with probability $\epsilon$. The value of $\epsilon$ was linearly decayed from an initial value of $0.5$ to a final minimum of $0.01$ over a specified fraction of the phase's total timesteps. For the continuous PointMaze environment, this decay occurred over the first 40\% of each 333,333-step phase. In the discrete MiniGrid environment, the decay occurred over the first 75\% of all phases. This forced stochasticity ensured that the agent encountered structurally diverse states, reliably triggering the GWR insertion threshold ($a_s < a_t$) and preventing premature convergence on suboptimal topological graphs.

Crucially, because this experimental setup evaluated performance across explicit, hard-coded task boundaries (A-B-A sequences), the $\epsilon$ value was manually reset to $0.5$ at the onset of each new phase to restimulate topological mapping. While this explicit reset isolated our evaluation of the architecture's capacity to resist catastrophic forgetting and achieve backward transfer, deploying ATLAS in a strictly boundary-agnostic, lifelong learning setting would require a dynamic exploration trigger. Theoretically, this could be seamlessly integrated by linking the $\epsilon$ value directly to the agent's reward prediction error or the GWR network's node-insertion rate. Under such a paradigm, the exploration rate would naturally spike when the agent encounters unexpected environmental dynamics or novel topologies, entirely bypassing the need for explicit task-boundary signals.

\section{Algorithmic Details}
\label{app:algorithms}

\subsection{MiniGrid Offline Planning Implementation}
While the ATLAS framework formally defines the high-level offline planning ("dreaming") phase via exact matrix inversion, early iterations of our discrete MiniGrid experiments utilized an iterative Bellman update to propagate values backward across the GWR graph. 

Because the iterative value propagation approach asymptotically converges to the exact analytical solution $\mathbf{\Psi} = (\mathbf{I} - \gamma \mathbf{T})^{-1}$, the resulting optimal policies extracted by the Breadth-First Search are functionally identical. Due to the bounded nature of the GWR graph in our tested environments, we transitioned to the exact analytical inversion for the continuous PointMaze experiments to maximize computational efficiency during the periodic offline updates. Future implementations scaled to massive, open-ended environments where cubic matrix inversion becomes computationally prohibitive may seamlessly revert to the iterative propagation method without loss of theoretical guarantees.
\clearpage
\section{Environment Configurations}
\label{app:configs}

\begin{figure}[h]
    \centering
    % First Subfigure
    \begin{subfigure}[t]{0.32\textwidth}
        \centering
        \includegraphics[width=\textwidth]{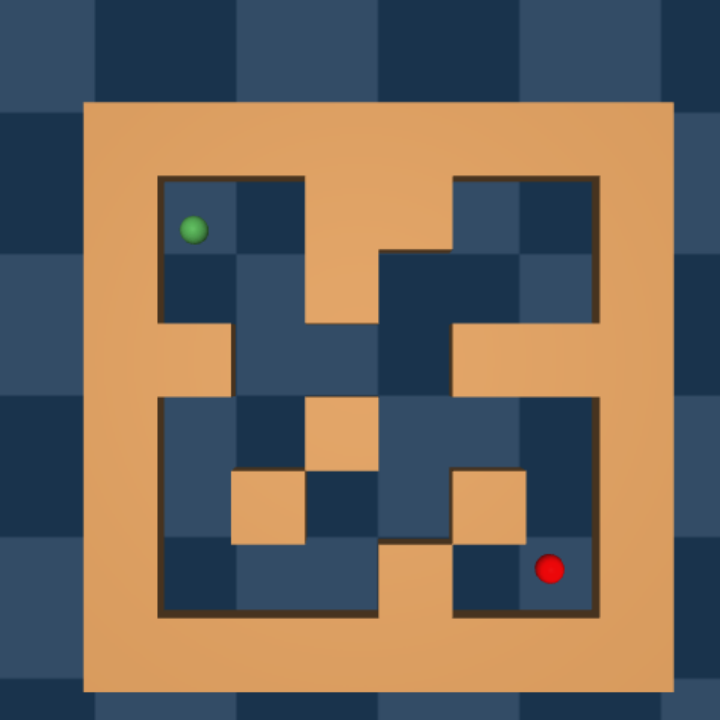}
        \caption{Phase A}
        \label{fig:pm_phase_a}
    \end{subfigure}
    \hfill
    % Second Subfigure
    \begin{subfigure}[t]{0.32\textwidth}
        \centering
        \includegraphics[width=\textwidth]{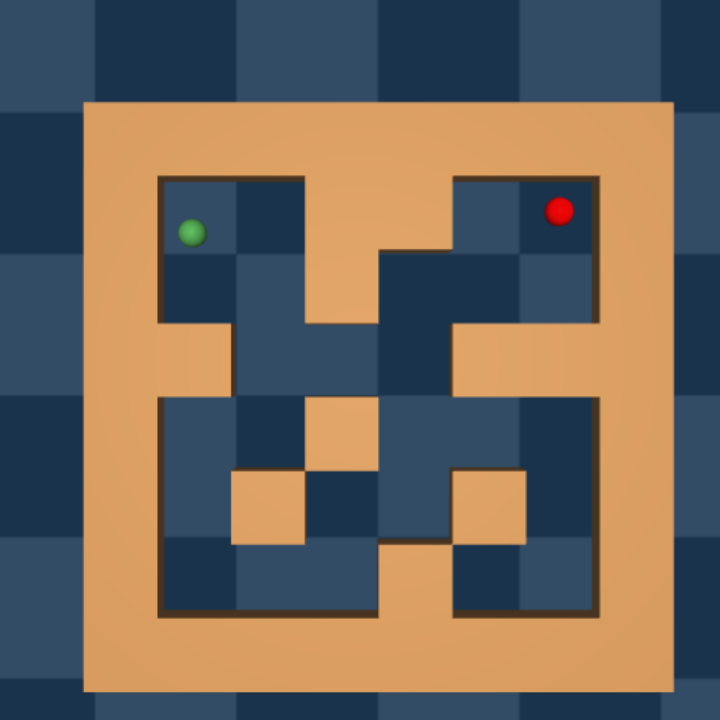}
        \caption{Phase B}
        \label{fig:pm_phase_b}
    \end{subfigure}
    \hfill
    % Third Subfigure
    \begin{subfigure}[t]{0.32\textwidth}
        \centering
        \includegraphics[width=\textwidth]{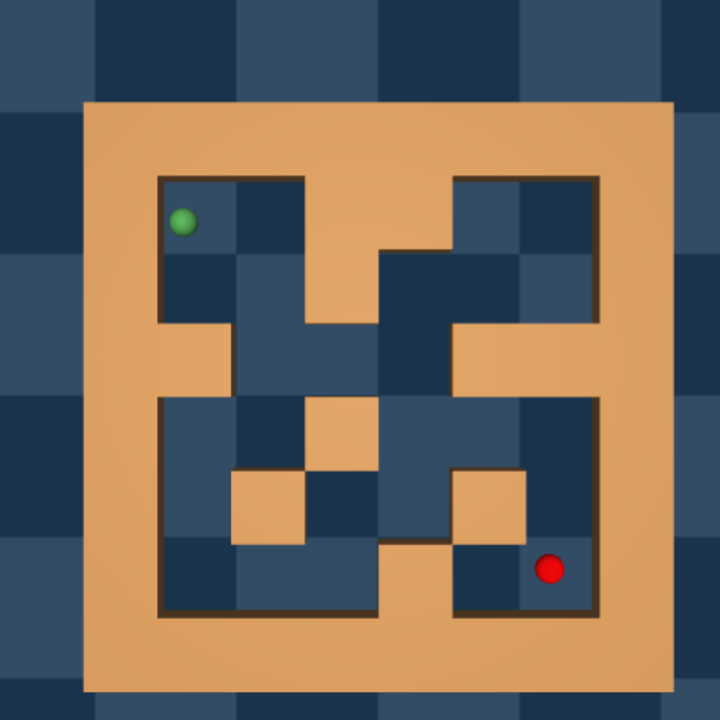}
        \caption{Phase A2}
        \label{fig:pm_phase_a2}
    \end{subfigure}
    
    \caption{Visual configuration of the PointMaze Medium experiment. In Phase A, the goal (red) is positioned in the bottom right, moved to the top left in Phase B, and returned to its original position in Phase A2. This sequence explicitly tests the agent's ability to handle shifting reward dynamics without topological collapse.}
    \label{fig:pointmaze_images}
\end{figure}

\begin{figure}[h]
    \centering
    % First Subfigure
    \begin{subfigure}[b]{0.32\textwidth}
        \centering
        \includegraphics[width=\textwidth]{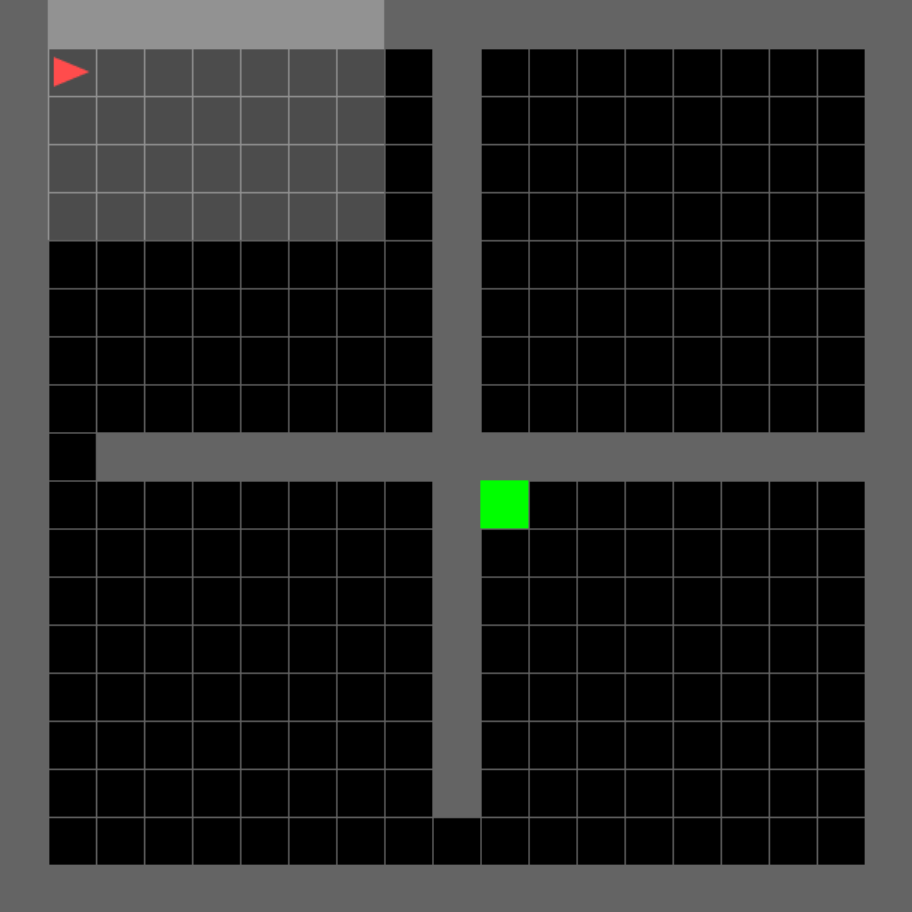}
        \caption{Phase A: Top-Right Room Locked}
        \label{fig:mg_phase_a}
    \end{subfigure}
    \hfill
    % Second Subfigure
    \begin{subfigure}[b]{0.32\textwidth}
        \centering
        \includegraphics[width=\textwidth]{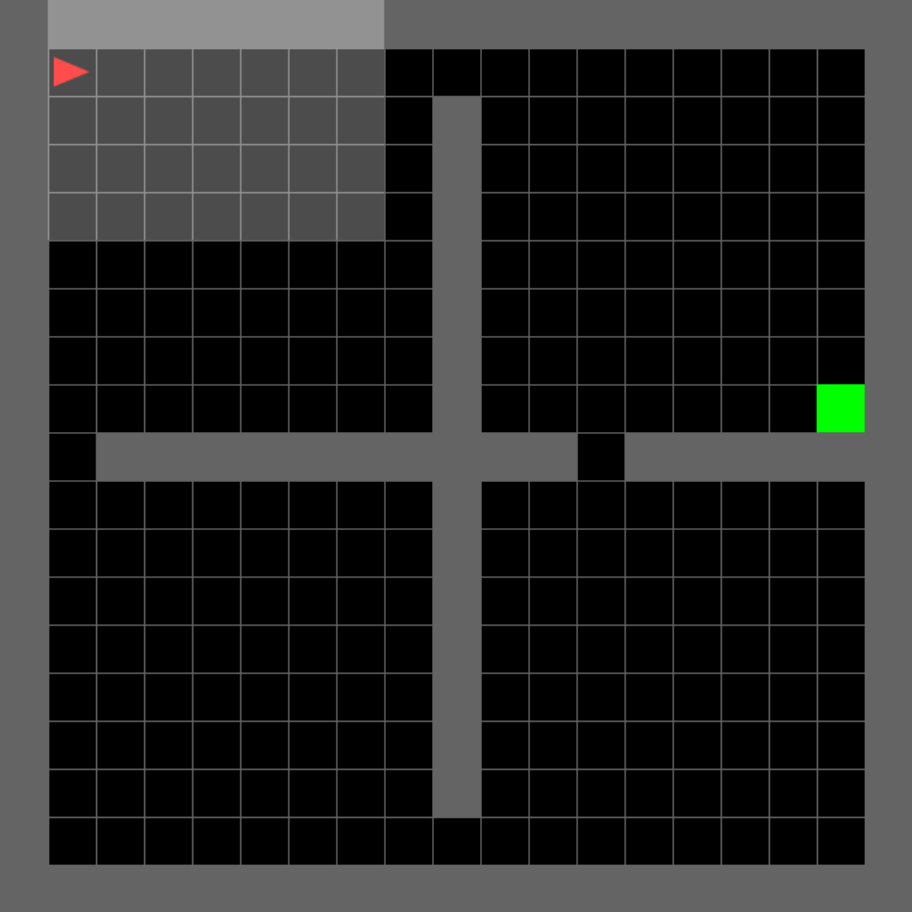}
        \caption{Phase B: Room Opened \& Goal Moved}
        \label{fig:mg_phase_b}
    \end{subfigure}
    \hfill
    % Third Subfigure
    \begin{subfigure}[b]{0.32\textwidth}
        \centering
        \includegraphics[width=\textwidth]{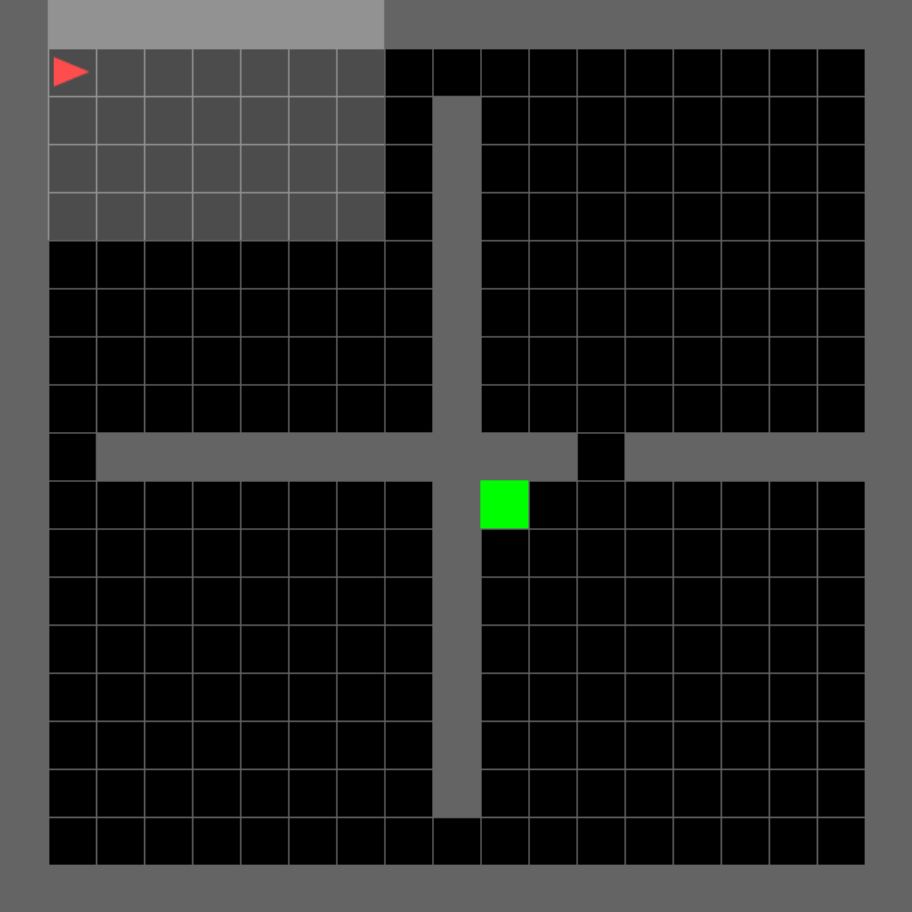}
        \caption{Phase A2: Room Open \& Goal Returned}
        \label{fig:mg_phase_a2}
    \end{subfigure}
    
    \caption{Design of the Locked Room MiniGrid experiment. The layout is structured such that unlocking the top-right room reveals a significantly shorter pathway to the objective. This evaluates the agent's ability to rapidly integrate new topological transitions and update its established policy.}
    \label{fig:minigrid_images}
\end{figure}

This appendix provides visual representations of the primary non-stationary environments utilized for evaluation: the continuous PointMaze Medium (Figure \ref{fig:pointmaze_images}) and the discrete MiniGrid Locked Room (Figure \ref{fig:minigrid_images}).

\vspace{1em} % Adds a little breathing room between the two large figures

\section{Additional Experiments}
\label{app:experiments}

\subsection{Minigrid Two-Phase Experiment}
\begin{figure}
    \centering
    \includegraphics[width=1\linewidth]{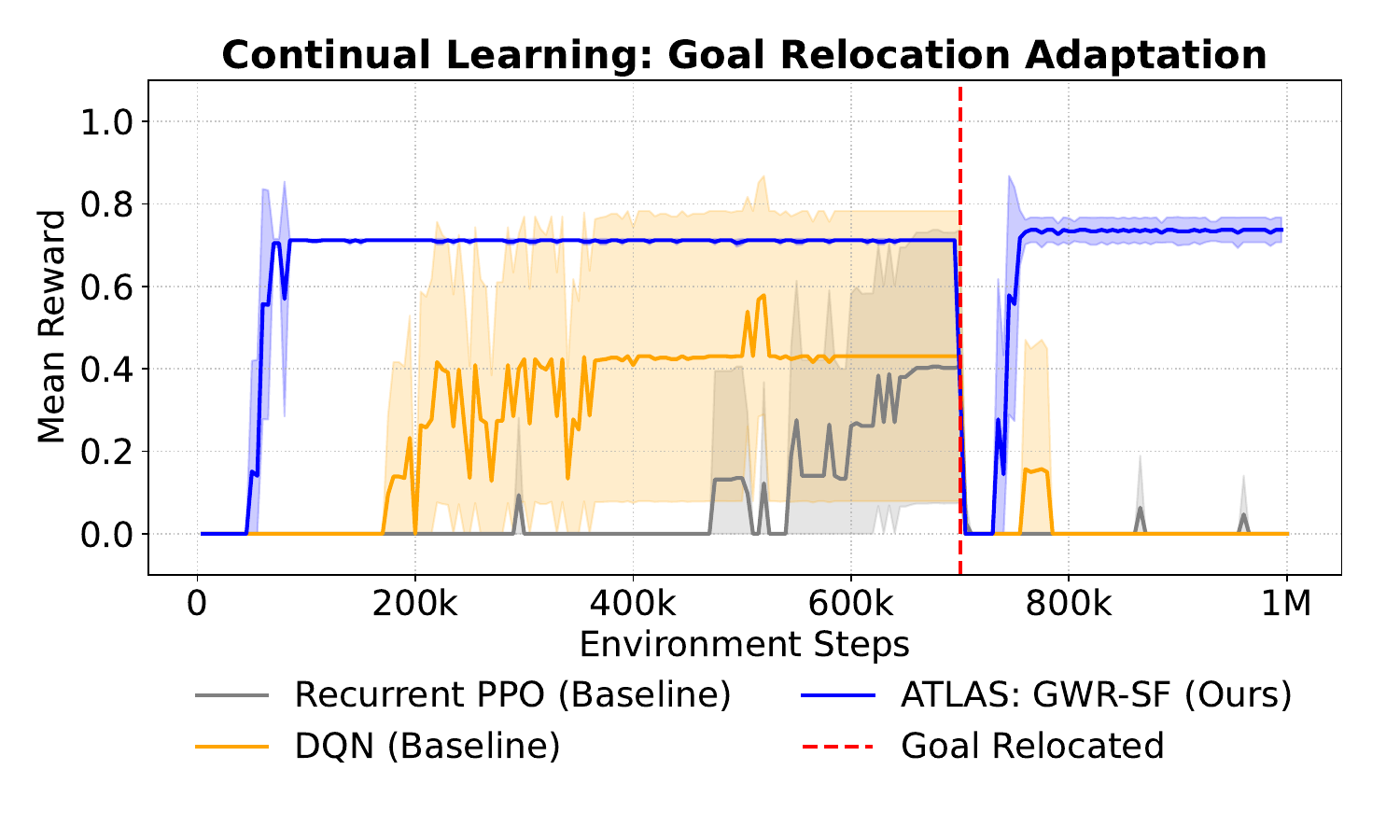}
    \caption{Continual learning performance during goal relocation. Mean episodic reward across $1 \times 10^6$ environment steps. At 700,000 steps (red dashed line), the navigation goal is abruptly relocated to a different room. While baseline methods (Recurrent PPO and DQN) suffer from catastrophic forgetting and fail to recover a viable policy, ATLAS rapidly adapts. By utilizing its GWR topology to protect prior structural knowledge, ATLAS successfully re-converges almost immediately with minimal variance (shaded regions represent $\pm$ one standard deviation across multiple seeds constrained between 0 and 1.)}
    \label{fig:baseline}
\end{figure}

In this test in MiniGrid FourRooms, we compared ATLAS against a PPO and DQN baseline. The goal was initially positioned at $(15, 15)$ for the first 700,000 timesteps. To test the algorithms' robustness to non-stationary reward dynamics, the goal was subsequently relocated to $(1, 17)$ for the remaining 300,000 timesteps. The comparative performance of ATLAS against the baseline models is illustrated in Figure \ref{fig:baseline}.

The empirical results demonstrate several key advantages of the ATLAS architecture. First, the narrow standard deviation bands indicate that ATLAS exhibits significantly higher consistency across random initializations compared to both DQN and Recurrent PPO. Across all evaluated seeds, ATLAS reliably converged on a highly performant and stable policy, avoiding the high algorithmic variance that plagued the baseline models. 

Second, ATLAS demonstrates superior sample efficiency. While enhanced sample efficiency is a theoretical hallmark of model-based reinforcement learning, it is notable that ATLAS achieves near-optimal performance well before either baseline model registers any meaningful reward. 

Crucially, the relocation of the goal at step 700,000 induces an immediate drop to zero reward across all agents. However, ATLAS rapidly recovers, adapting its policy to the new goal location almost instantaneously without forgetting the underlying environmental topology. In contrast, DQN and PPO experience catastrophic forgetting; their performance collapses, and their model-free architectures must expensively relearn the environment. While the baselines show early signs of potential recovery, achieving parity with ATLAS would necessitate hundreds of thousands of additional environment interactions. 

Ultimately, these results validate that by decoupling the transition dynamics from the reward signal using a GWR topology and Successor Features, ATLAS successfully mitigates catastrophic forgetting while delivering marked improvements in consistency and sample efficiency over standard model-free approaches.

\subsection{Minigrid Three-Phase Experiment}

\begin{figure}[htbp]
    \centering
    \includegraphics[width=1\linewidth]{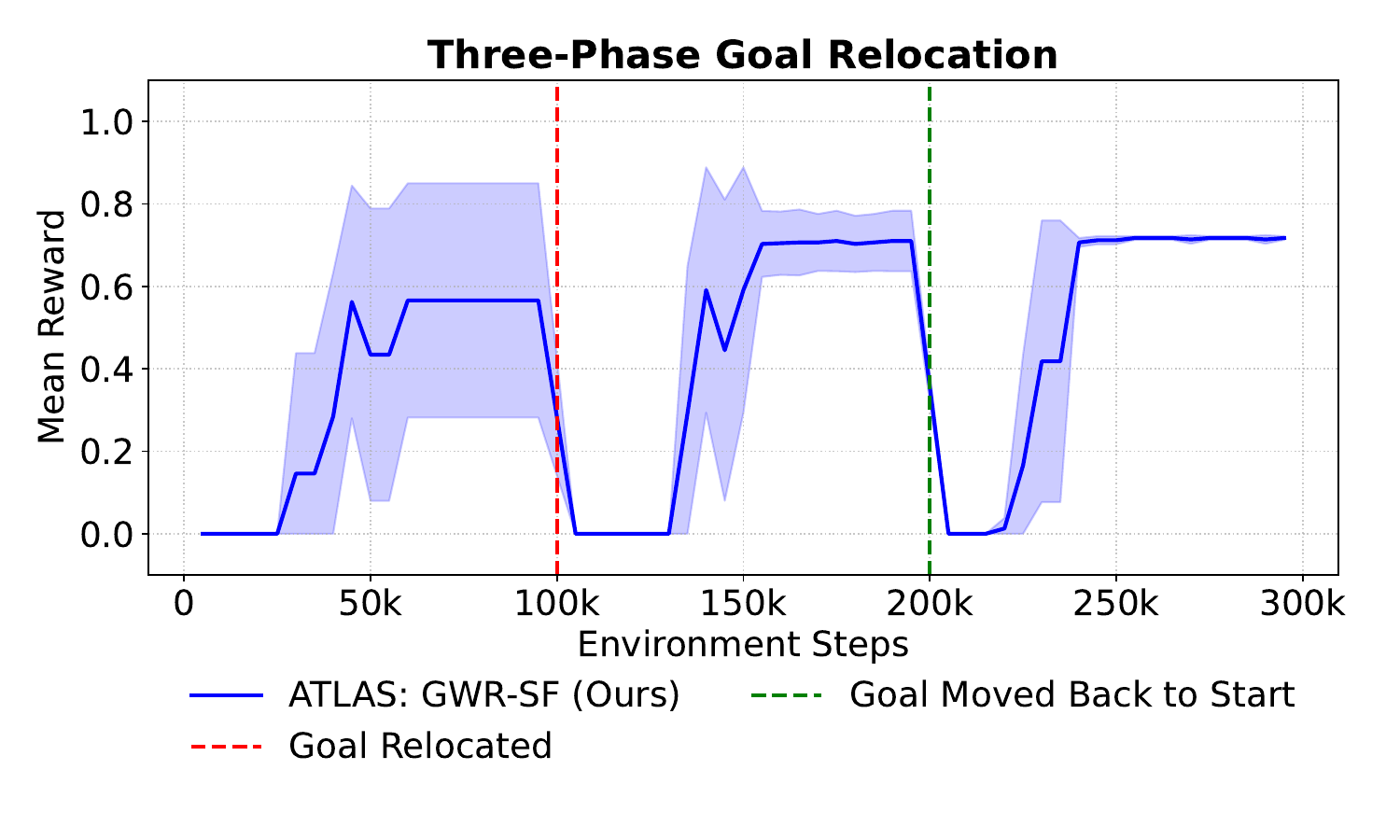}
    \caption{Mean reward across 5 random seeds during the A-B-A goal relocation experiment. Vertical red dashed lines at steps 100,000 and 200,000 indicate structural task shifts. ATLAS demonstrates positive backward transfer, achieving a more stable and higher-reward policy upon returning to the initial task.}
    \label{fig:three_phase}
\end{figure}

In our secondary evaluation, we subjected the agent to an A-B-A task sequence to evaluate its capacity for continual learning and knowledge transfer. The goal was positioned at $(15, 15)$ for the first 100,000 timesteps (Task A), relocated to $(1, 17)$ for the subsequent 100,000 timesteps (Task B), and finally returned to $(15, 15)$ for the final 100,000 timesteps (Task A). Given the baseline models' inability to recover from a single goal shift in the previous experiment, this evaluation isolates the ATLAS architecture to assess its specific relearning dynamics. The results are presented in Figure \ref{fig:three_phase}. 

The empirical data highlights several distinct phases of the agent's topological maturation. During the initial 100,000 timesteps, the agent successfully located the goal across all seeds; however, the resulting policy exhibited high variance and relatively low asymptotic reward. This instability is a natural consequence of the GWR network actively initializing and expanding its topological map from a blank state.

Upon the first relocation at step 100,000, ATLAS leveraged this partially constructed environmental knowledge, requiring only 25,000 steps to consistently locate the new goal. Notably, the policy formed during this second phase exhibited significantly reduced variance. 

Crucially, when the goal was reverted to its initial location at step 200,000, ATLAS required only 20,000 steps to locate the target, an improvement over the initial discovery phase. More importantly, the recovered policy in this third phase significantly outperformed the policy learned during the first 100,000 steps, achieving higher mean returns with near-zero variance. 

Overall, this experiment demonstrates that ATLAS exhibits positive backward transfer. Rather than simply memorizing isolated trajectories, the architecture utilizes the comprehensive spatial and topological knowledge acquired during subsequent exploration to refine and optimize its approach to previously encountered tasks.

\end{document}